\documentclass[sigconf,nonacm]{acmart}

\setcopyright{none}
\renewcommand\footnotetextcopyrightpermission[1]{}

\usepackage{amsmath}
\usepackage{booktabs}
\usepackage{multirow}
\usepackage{algorithmicx}
\usepackage{algpseudocode}
\usepackage{graphicx}
\usepackage{xspace}
\usepackage{dblfloatfix}
\usepackage{balance}
\usepackage{placeins}
\usepackage{ragged2e}
\usepackage{float}

\graphicspath{{./}}

\newcommand{\statesight}{\textsc{StateSight}\xspace}
\newcommand{\statesightsteps}{\textsc{StateSight-Steps}\xspace}

\algrenewcommand\alglinenumber[1]{\footnotesize #1:}
\algrenewcommand\algorithmicindent{1.15em}

\title{StateSight: Benchmarking Latent Spatial-State Reconstruction in Vision-Language Models}

\author{Michelle Lin}
\affiliation{
  \institution{Thomas Jefferson High School for Science and Technology}
  \city{Alexandria}
  \state{Virginia}
  \country{USA}
}
\email{2028mlin@tjhsst.edu}

\begin{document}

\begin{abstract}
Vision-language models are increasingly used for multimodal question answering, yet their ability to reconstruct latent spatial structure from a single image remains difficult to isolate. Broad benchmarks often combine perception, optical character recognition, domain knowledge, linguistic priors, and reasoning in the same evaluation. We introduce \statesight, a procedurally generated benchmark for cube-net opposite-face reasoning, occluded cube-tower counting, and 4-neighbor connected-component counting. Each task family contains 300 single-image prompts with deterministic oracle labels and exact-match scoring. OpenAI GPT-5.5, using the API model identifier gpt-5.5, achieved 59.3\%, 33.3\%, and 28.3\% accuracy across the three tasks, while Claude Sonnet 5 achieved 53.3\%, 18.7\%, and 7.3\%. All final direct runs had zero format errors. A 30-participant human baseline on 60 items exceeded both models on every task, with mean accuracies of 80.8\%, 68.8\%, and 64.3\%. Visible-derivation analysis identified recurring errors in image-state reconstruction and reasoning procedure. We also introduce \statesightsteps, a companion dataset of 900 interleaved image-text examples and 3,600 deterministic intermediate visual states. The results show that format-valid responses can mask failures to recover the spatial structure required for verifiable visual inference.
\end{abstract}

\keywords{vision-language models, visual reasoning, spatial reasoning, multimodal benchmarks, procedural generation, latent state}

\maketitle

\section{Introduction}

Vision-language models have shown strong performance on image captioning, visual question answering, document understanding, and multimodal instruction following. These results have encouraged the view that modern systems possess increasingly general visual reasoning abilities. However, benchmark accuracy alone may not reveal whether a system has recovered the internal structure needed to solve a problem. In many multimodal evaluations, object recognition, optical character recognition, language modeling, domain knowledge, and reasoning contribute to a single final answer. A model can therefore succeed through several pathways, including the use of regularities that do not require the intended visual inference.

Diagnostic benchmarks attempt to separate these pathways. CLEVR uses generated scenes and detailed functional annotations to expose weaknesses that can remain hidden in natural-image visual question answering \cite{johnson2017clevr}. GQA extends structured reasoning evaluation to real-world scenes through scene graphs and compositional programs \cite{hudson2019gqa}. More recent benchmarks, including BLINK and Mind the Gap, show that multimodal models continue to struggle with core perception and spatial reasoning even when they perform well on broader evaluations \cite{fu2024blink,stogiannidis2025mind}. These findings motivate a narrower question: can a vision-language model reconstruct the latent spatial state that generated a clean visual prompt?

This work studies that question through \statesight. The benchmark removes natural-image semantics and focuses on three spatial reconstruction problems. The first requires mapping a two-dimensional cube net to the three-dimensional relation between opposite faces. The second requires recovering hidden cube occupancy from an isometric rendering generated from a height map. The third requires converting a binary grid into its 4-neighbor connected-component topology. Each answer is computed directly from the generator state. This makes it possible to distinguish output-format compliance from task correctness and to analyze errors against an exact answer oracle.

The benchmark contains 900 items, with 300 examples for each task family. The evaluated systems were OpenAI GPT-5.5, using the API model identifier gpt-5.5, and Claude Sonnet 5. Their responses were normalized to the required letter or positive integer and compared against the corresponding oracle label. GPT-5.5 outperformed Claude Sonnet 5 across all three direct tasks, but both models declined substantially on hidden occupancy and connected-component counting. Every final direct run contained zero format errors.

A human baseline tested whether the benchmark items were interpretable under comparable answer-only conditions. Thirty adult participants completed a 60-item packet containing 20 examples from each task. Human mean accuracy remained above both models on all three tasks, although performance was not perfect on towers or grids.

The project also examines visible derivations and intermediate visual states. Models were separately asked to return an answer, a derivation, and an explanation on matched 30-item subsets. These outputs are visible response traces, not hidden chain-of-thought. Conservative coding of the traces found frequent errors in perception or reconstruction of the input state, followed by errors in reasoning procedure. Motivated by work on visual chain-of-thought and multimodal world models \cite{zhou2025mira,wu2026visual}, the project further introduces \statesightsteps. This companion layer provides deterministic intermediate images aligned with step-level text, allowing future experiments to test whether explicit visual-state supervision improves performance.

The central contribution is twofold. \statesight provides a diagnostic benchmark for latent spatial-state reconstruction, while \statesightsteps provides generator-defined targets for studying intermediate visual supervision. The evaluation offers controlled evidence that two strong vision-language models can produce valid answers while failing to recover the spatial representation required by the task.

\begin{figure*}[!t]
    \centering
    \includegraphics[
        width=0.92\textwidth,
        keepaspectratio
    ]{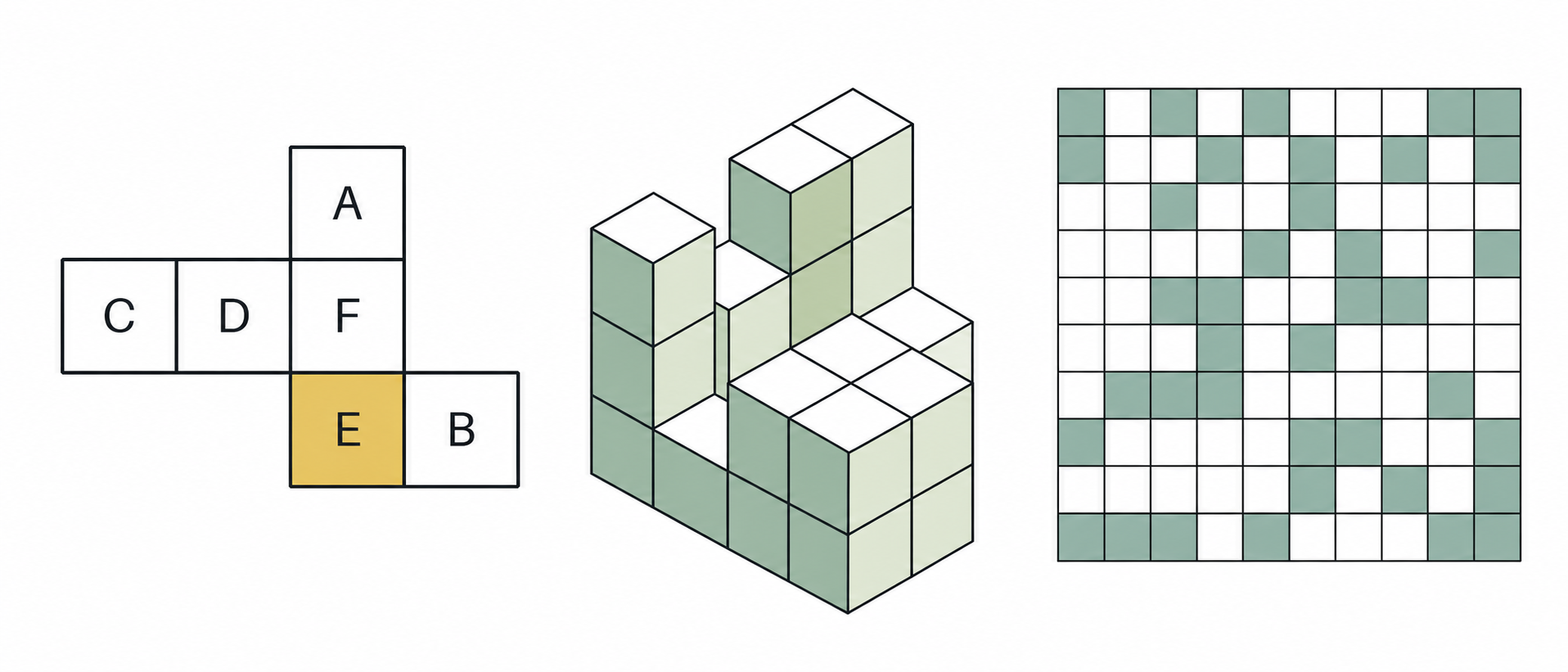}
    \caption{\textbf{The three \statesight task families: cube-net folding, hidden cube occupancy, and connected-component structure.} \normalfont The panels isolate planar-to-three-dimensional face relations, occluded occupancy, and 4-neighbor topology under one controlled rendering style.}
    \label{fig:families}
\end{figure*}

\section{Related Works}

\subsection{Visual Reasoning Benchmarks}

Visual question answering benchmarks have long been used to measure whether systems can combine image understanding with language. Early datasets revealed that models could exploit linguistic and answer-distribution biases rather than reason about visual content. CLEVR addressed this problem through generated scenes, controlled question templates, and functional programs that identify the reasoning steps associated with each answer \cite{johnson2017clevr}. GQA used scene graphs and compositional question generation to provide similar structure in real-world imagery \cite{hudson2019gqa}. These benchmarks established the value of generator-controlled data for identifying specific model weaknesses.

Large multimodal benchmarks have expanded evaluation across mathematics, diagrams, charts, academic disciplines, and heterogeneous image formats. MathVista emphasizes mathematical reasoning in visual contexts, while MMMU measures broad expert-level multimodal understanding \cite{lu2023mathvista,yue2023mmmu}. Beyond the diagnostic motivation stated in the introduction, \statesight differs methodologically by fixing the complete data-generating process: rendering style, answer type, target relation, and oracle state. This control permits item-level reconstruction errors to be analyzed directly rather than inferred from a broad aggregate score.

\subsection{Compositionality, Perception, and Spatial Reasoning}

Winoground demonstrates that strong vision-language systems can fail compositional tests even when images and captions use familiar objects and words \cite{thrush2022winoground}. BLINK similarly shows that multimodal LLMs perform poorly on core visual perception tasks that humans solve accurately \cite{fu2024blink}. Perception Test extends diagnostic evaluation to multimodal video understanding and reports a large gap between human and model performance \cite{patraucean2023perception}. Together, these results suggest that broad benchmark success does not guarantee reliable performance on focused visual capabilities.

Spatial understanding has also become a direct target of evaluation. SpatialBot introduces spatial question answering and depth-aware model inputs \cite{cai2024spatialbot}. Mind the Gap analyzes spatial relations, orientation, navigation, mental rotation, and spatial visualization across generated and natural images \cite{stogiannidis2025mind}. \statesight is narrower than these benchmarks. It does not attempt to represent the full range of human spatial cognition. Instead, it focuses on three generated tasks for which the underlying spatial state and final label are fully specified by code.

\subsection{Intermediate Visual Reasoning}

Textual chain-of-thought can help models decompose problems, but some spatial transformations are difficult to represent through language alone. MIRA studies problems in which intermediate images such as sketches, diagrams, and paths are useful or necessary for successful reasoning \cite{zhou2025mira}. Work on multimodal world models argues that visual generation can provide a more suitable reasoning substrate for tasks grounded in physical and spatial structure \cite{wu2026visual}. These studies motivate the \statesightsteps extension. Its intermediate images are generated from task metadata rather than predicted by a model, which makes each visual state machine-checkable. The resulting dataset can be used to test whether supplying or learning these states improves final-answer accuracy.

\par\vspace{-0.9em}
\section{Methodologies}
\vspace{-0.15em}

\subsection{Implementation Details}

The benchmark was implemented as a programmatic generation and evaluation pipeline. Each task family defines a latent state space, renderer, answer oracle, prompt template, and parser. A sampled state produces both the rendered image and its label, keeping the prompt and ground truth aligned by construction. Task 1 accepts one of the six face labels \texttt{A} through \texttt{F}. Task 2 returns a positive integer equal to the total number of unit cubes, including hidden supports; Task 3 returns a positive integer equal to the number of 4-neighbor connected gray regions.

OpenAI GPT-5.5 was evaluated using the API model identifier gpt-5.5 with reasoning effort disabled, low text verbosity, low image detail, and a 128-token output limit. Claude Sonnet 5 was evaluated with thinking disabled, low effort, and a 64-token output limit. Each model received 300 items per task under a fixed answer-only prompt. Local validation checked row counts, duplicate identifiers, missing responses, API errors, parsing failures, and exact-match scores. Every final direct file contained 300 completed rows and zero format errors.

For each model-task pair, the observed proportion correct was summarized with a Wilson 95\% confidence interval over 300 binary outcomes. Differences between GPT-5.5 and Claude Sonnet 5 were estimated by bootstrap resampling shared item identifiers, preserving the paired evaluation design. Human intervals were calculated from the 30 participant-level task accuracies. Because the human packet contained 20 items per task whereas the direct model evaluation contained 300, human-model gaps are reported as descriptive benchmark-level comparisons rather than item-matched significance tests.

\vspace{0.6em}
\begin{figure*}[!t]
    \centering
    \includegraphics[
        width=0.88\textwidth,
        keepaspectratio
    ]{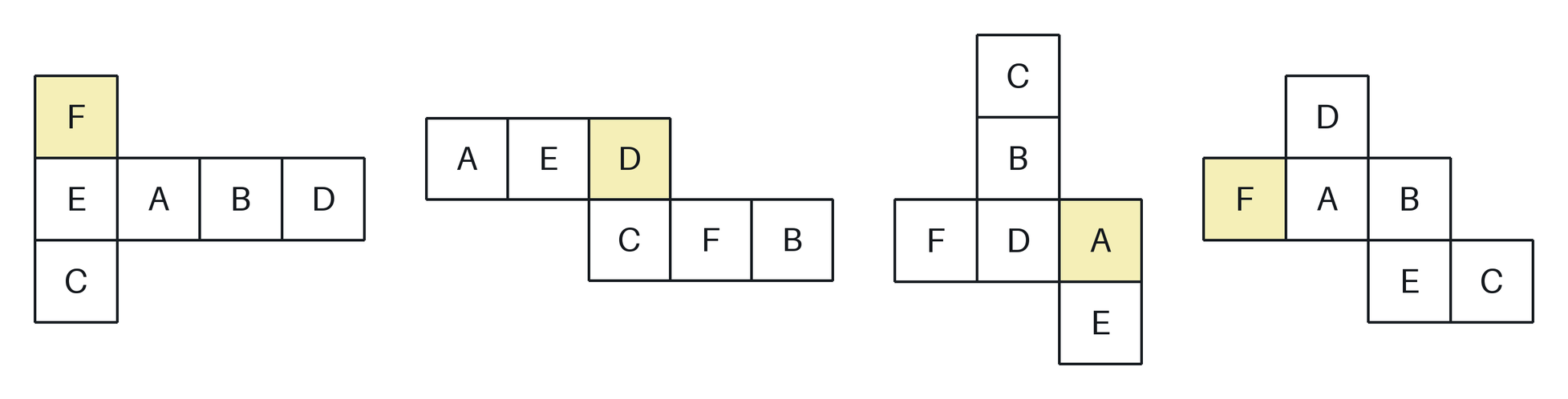}
    \caption{\textbf{Four generator-defined intermediate visual states from one \statesightsteps example.} \normalfont The panels share one symbolic instance and appear in generator-defined order, making each intermediate state independently checkable.}
    \label{fig:latent}
\end{figure*}

\par\vspace{0.55em}

\noindent
\begin{minipage}{\columnwidth}

\begin{minipage}{0.98\columnwidth}

\hrule height 0.6pt
\vspace{0.25em}

\noindent\textbf{Algorithm 1} Latent-State Evaluation

\vspace{0.18em}
\hrule height 0.35pt
\vspace{0.3em}

{\footnotesize
\begin{algorithmic}[1]

\State \textbf{Inputs:} task set $\mathcal{T}$, model set $\mathcal{M}$, item count $N$
\State For each task $t$, define renderer $R_t$, oracle $F_t$, parser $\pi_t$

\For{each task $t\in\mathcal{T}$}
    \For{$i=1,\ldots,N$}
        \State Draw latent state $z_i$
        \State Form image-label pair
        \Statex \hspace{\algorithmicindent}
        $(x_i,y_i)\leftarrow(R_t(z_i),F_t(z_i))$
    \EndFor

    \For{each model $M\in\mathcal{M}$}
        \State Compute
        \Statex \hspace{\algorithmicindent}
        $\displaystyle
        \mathrm{EM}_{M,t}
        \leftarrow
        \frac{1}{N}
        \sum_{i=1}^{N}
        \mathbf{1}
        \left[
        \pi_t(M(x_i,p_t))=y_i
        \right]$
        \State Record accuracy and format validity
    \EndFor
\EndFor

\State \textbf{return} task-wise evaluation results

\end{algorithmic}
}

\vspace{0.25em}
\hrule height 0.6pt

\end{minipage}

\par\vspace{0.28em}

\noindent
The renderer and oracle share one latent state; parsing only
normalizes the returned label or count and awards no partial credit.

\end{minipage}

\par\vspace{0.65em}

\subsection{Dataset Description}

The problem considered in this work is latent spatial-state reconstruction from a single rendered image. \statesight contains 900 benchmark items, divided evenly across three task families. The image set is generated by custom task-specific renderers rather than collected from external sources. Every image is paired with state metadata and a deterministic oracle label. This design supports reproducible regeneration, exact scoring, and controlled analysis of task properties.

Task 1 contains cube nets. A valid cube-net layout is generated, faces are assigned letter labels, and one face is highlighted. The oracle computes which labeled face lies opposite the highlighted face after folding. The rendered image contains the net, labels, and highlight but does not show the folded cube. The task therefore requires a two-dimensional to three-dimensional transformation.

Task 2 contains isometric block towers generated from integer height maps. The height map defines the number of unit cubes at each occupied footprint location. The renderer exposes the visible surfaces while the oracle sums all column heights, including hidden supporting cubes. These items test whether the model can reconstruct occluded occupancy rather than count visible faces or top surfaces.

Task 3 contains fixed 10 by 10 binary grids. Gray cells are connected only through shared edges. The oracle applies 4-neighbor connected-component labeling and returns the number of gray regions. The model must preserve local adjacency while tracking component merges across the full image. Diagonal contact does not create connectivity.

A companion dataset, \statesightsteps, provides intermediate states for each benchmark item. Cube-net examples expose the relevant net coordinates, folded directions, and opposite-face relation. Tower examples provide the height map, occupied footprint, and layer slices. Grid examples provide the binary mask, component map, and numbered final grouping. The full layer contains 900 examples and 3,600 intermediate images. These traces are deterministic and are intended for future prompting, supervision, or fine-tuning experiments; the present study does not claim that training on them improves performance.

\subsection{Human Baseline}

Thirty adult participants completed the same 60-item packet, containing 20 items per task. No fixed time limit was imposed, and completion time was not used in scoring. Image zooming was allowed, while calculators, AI systems, internet search, collaboration, and scratch drawings were prohibited. Participants completed the packet on laptops and reported that they had not previously seen the tasks.

Human responses were scored with the same exact-match rules as model responses. Task 1 answers were normalized to one of the six face labels, and Tasks 2 and 3 required positive-integer matches. Blank or invalid answers counted as incorrect. Participant-level task accuracy was used for statistical summaries. The convenience sample tests whether the benchmark is interpretable under the stated protocol rather than estimating population-level ability.

\subsection{Visible-Derivation and Item Analysis}

A separate visible-derivation condition requested a final answer, derivation, and explanation for 30 matched examples per task and model. These are observable responses, not hidden chain-of-thought. Results are reported separately because both the sample size and prompt differ from the 300-item direct condition.

The visible responses were coded into conservative categories based on their content. The largest category was wrong perception or reconstruction of the input image, followed by an incorrect reasoning approach. A small number contained fabricated visual facts, and some combined multiple error types. Automatic coding is treated as a diagnostic aid rather than causal ground truth.

Item-level analysis combines human and model outcomes on the 60-item human subset. Human-easy items have participant accuracy of at least 0.80, while human-hard items have accuracy below 0.70. Model outcomes are separated according to whether zero, one, or both evaluated models answered correctly. This non-overlapping taxonomy identifies items that are easy for humans but difficult for the models, while also isolating items that are difficult for both groups.

\subsection{Results and Discussion}

Table~\ref{tab:direct} reports direct answer-only accuracy. GPT-5.5 achieved 59.3\%, 33.3\%, and 28.3\% across Tasks 1--3, while Claude Sonnet 5 achieved 53.3\%, 18.7\%, and 7.3\%. The paired GPT-5.5-minus-Claude differences were 6.0, 14.7, and 21.0 percentage points; the corresponding bootstrap intervals excluded zero.

\begin{table}[!t]
\centering
\caption{Direct exact-match results over 300 items per task.}
\label{tab:direct}
\begin{tabular}{llrr}
\toprule
Task & Model & Correct & Accuracy \\
\midrule
\multirow{2}{*}{Cube nets}
& GPT-5.5 & 178/300 & 59.3\% \\
& Claude Sonnet 5 & 160/300 & 53.3\% \\
\midrule
\multirow{2}{*}{Cube towers}
& GPT-5.5 & 100/300 & 33.3\% \\
& Claude Sonnet 5 & 56/300 & 18.7\% \\
\midrule
\multirow{2}{*}{Connected grids}
& GPT-5.5 & 85/300 & 28.3\% \\
& Claude Sonnet 5 & 22/300 & 7.3\% \\
\bottomrule
\end{tabular}
\end{table}

The human baseline averaged 80.8\%, 68.8\%, and 64.3\% across the three tasks. Relative to GPT-5.5, the human-model gaps were 21.5, 35.5, and 36.0 percentage points; relative to Claude Sonnet 5, they were 27.5, 50.2, and 57.0 points. Human performance declined across tasks, but substantially less than model performance.

The visible-derivation condition did not consistently improve accuracy. GPT-5.5 achieved 50.0\%, 30.0\%, and 23.3\%, while Claude Sonnet 5 achieved 63.3\%, 16.7\%, and 30.0\%. Claude's higher values on Tasks 1 and 3 should not be interpreted as a direct prompting gain: this condition used only 30 matched items and a different prompt, whereas the direct condition used all 300 items. The difference can therefore reflect subset composition and sampling variability. Tower responses often omitted hidden supports, while grid responses sometimes stated the correct adjacency rule but still merged or split components incorrectly.

Across 180 visible-derivation responses, 72 were coded as incorrect image reconstruction, 41 as an incorrect reasoning procedure, 2 as mixed, and 1 as containing a fabricated visual fact; 64 were correct. These categories describe visible response content rather than the model's internal causal process.

The item-level taxonomy identified 5 human-easy items missed by both models, 6 solved by one model, and 8 solved by both. Among human-hard items, 16 were missed by both models, 4 were solved by one, and 2 were solved by both; 19 items fell in the middle human-accuracy range. The results do not imply that current models fail at all spatial reasoning: some items in every family were solved by both systems. They instead expose a consistent weakness in reconstructing latent spatial structure from clean images, even when answer formatting is correct.

\begin{figure}[H]
    \centering
    \includegraphics[width=0.92\columnwidth,keepaspectratio]{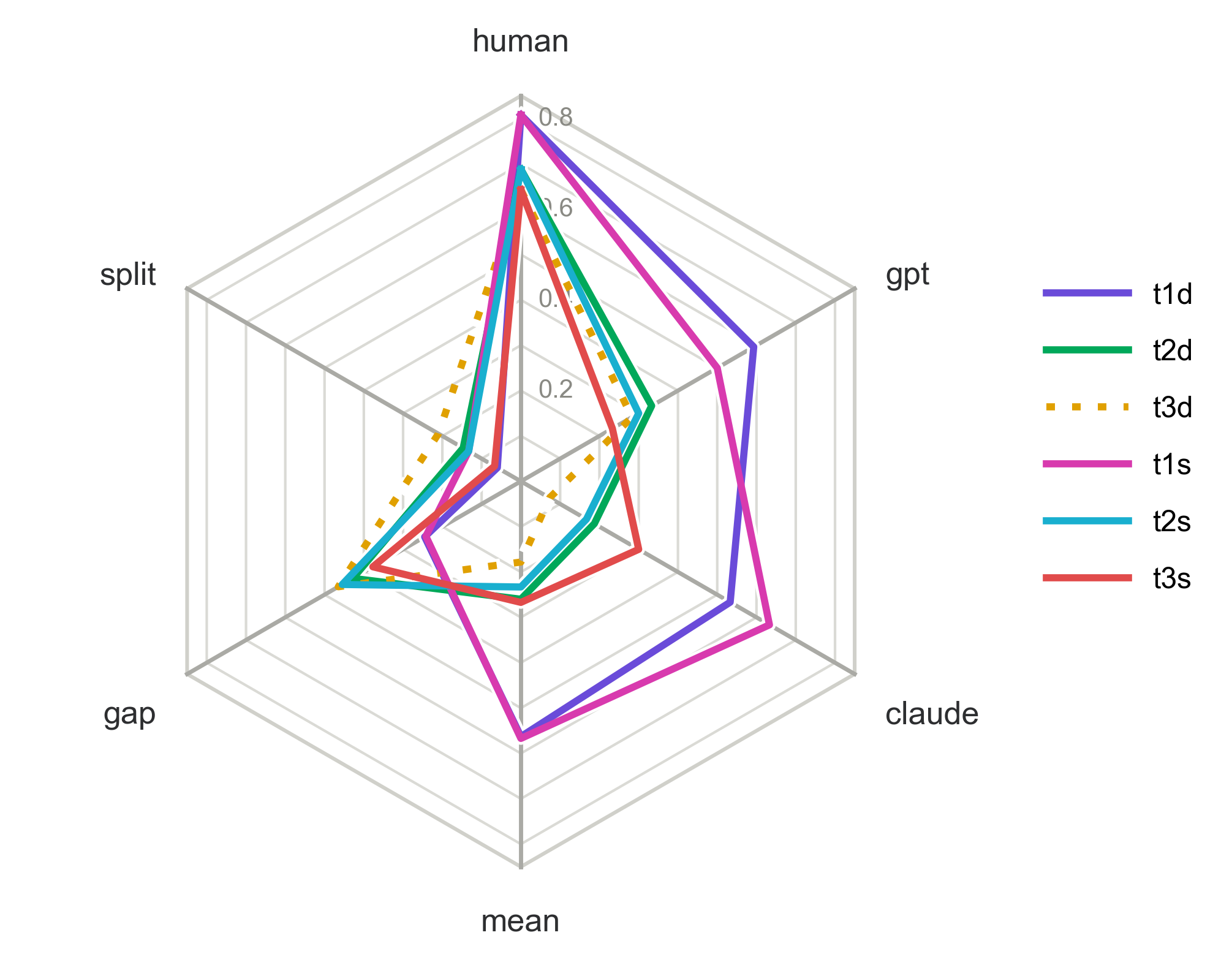}
    \caption{\textbf{Direct and visible-derivation performance profiles.} \normalfont Traces \texttt{t1d}--\texttt{t3d} denote direct conditions and \texttt{t1s}--\texttt{t3s} denote derivation subsets. The mixed accuracy, gap, and disagreement axes make this a descriptive synthesis rather than a statistical test.}
    \label{fig:radar}
\end{figure}

\clearpage
\onecolumn

\begin{center}
\includegraphics[width=0.84\textwidth,keepaspectratio]{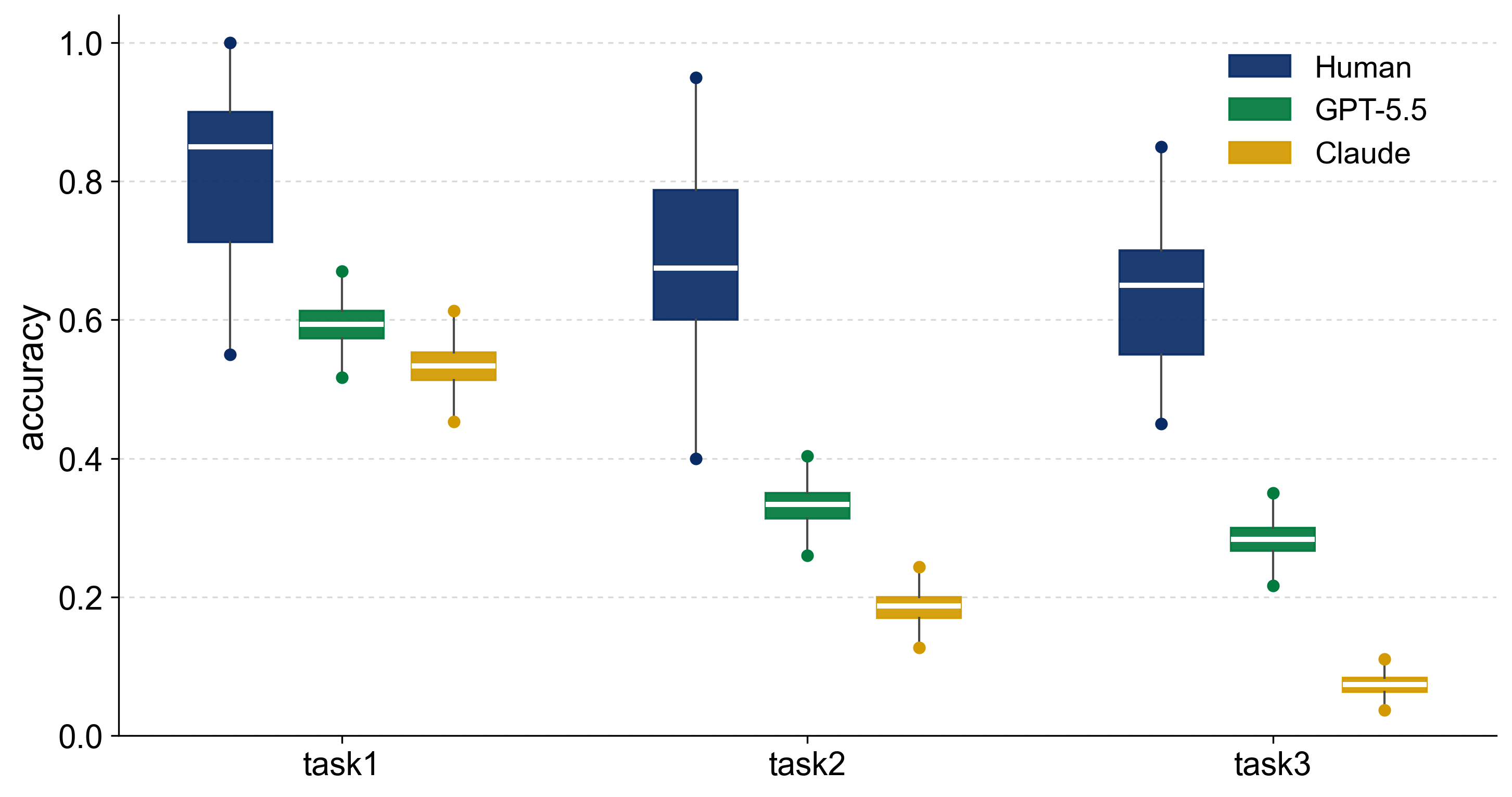}
\refstepcounter{figure}
\label{fig:boxplot}
\par\vspace{0.14em}
\noindent\textbf{Figure \thefigure: Human participant and model bootstrap distributions across the three tasks.}
\end{center}
\par\vspace{0.08em}
\noindent
Human boxes contain the 30 participant-level accuracies for each task. Model boxes contain 3,000 bootstrap resamples of the 300 item outcomes, not repeated model runs; the white line marks the median.

\vspace{0.45em}
\noindent
\begin{minipage}[t][0.47\textheight][s]{0.46\textwidth}
\vspace{0pt}
\centering
\makebox[\linewidth][c]{%
\includegraphics[width=1.10\linewidth,keepaspectratio]{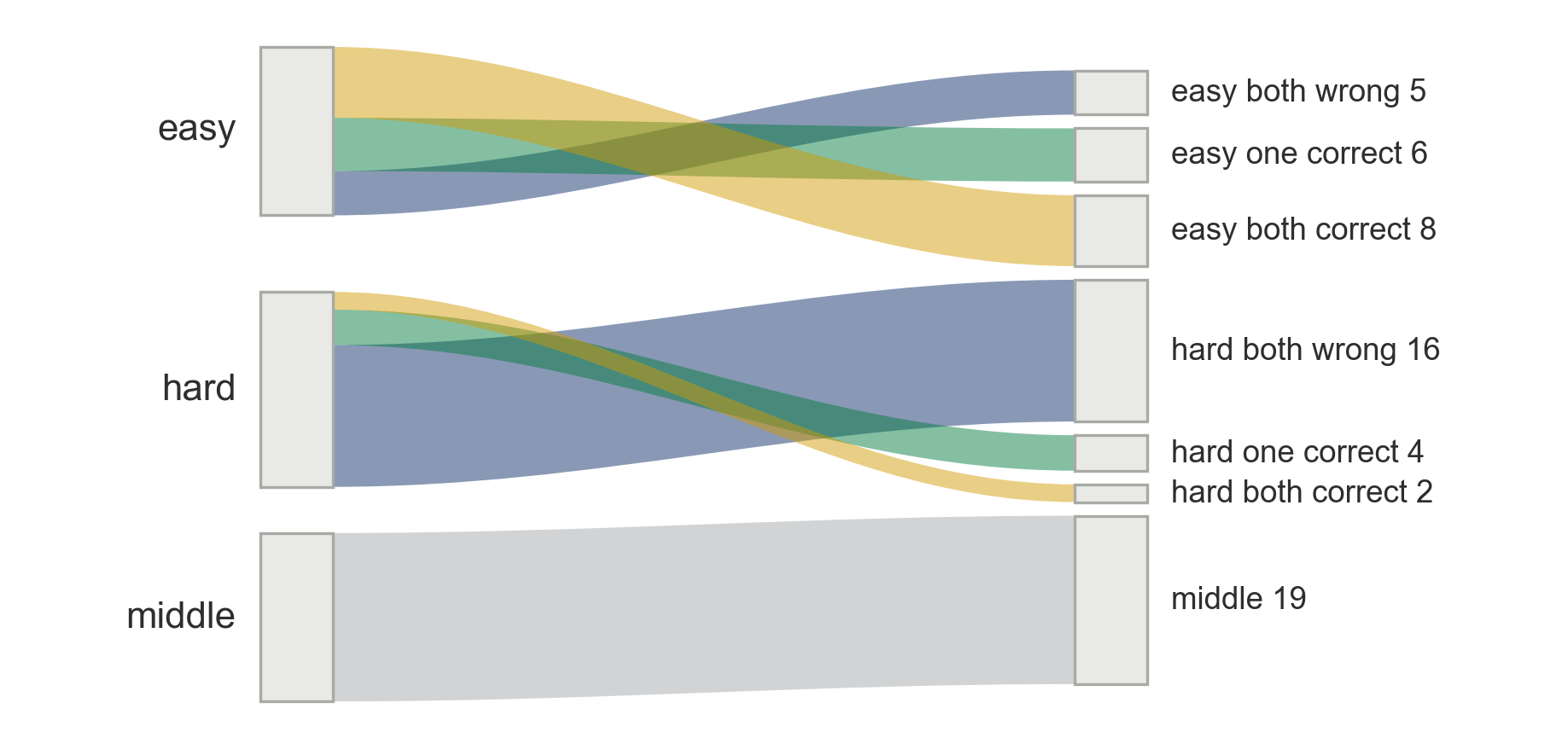}%
}
\refstepcounter{figure}
\label{fig:alluvial}
\par\vspace{0.10em}
\noindent\textbf{Figure \thefigure: Item-level human-difficulty and model-outcome categories.}
\par\vspace{0.06em}
\raggedright
Left nodes group the 60 human-baseline items as easy, hard, or middle by participant accuracy. Right nodes indicate whether neither model, one model, or both models answered correctly, and each flow's width equals its item count. The human-easy flows into model-error categories isolate cases that were broadly interpretable to participants but difficult for the evaluated systems.

\par\vspace{0.12em}
\noindent
The item-level view complements aggregate accuracy by separating model-specific failures from items that were difficult for humans as well. The five human-easy items missed by both models provide the strongest diagnostic subset for future reconstruction-level analysis.
\end{minipage}
\hfill
\begin{minipage}[t][0.47\textheight][s]{0.49\textwidth}
\vspace{0pt}
\section{Conclusions and Future Work}

In \statesight, two strong API models produced format-valid responses but remained below a 30-participant human baseline. The largest deficits occurred on hidden cube occupancy and connected-component topology. Visible derivations further showed that plausible explanations can accompany an incorrect reconstruction of the image state.

\vfill
The evaluation is limited to two proprietary models, a convenience-sample human baseline, and three deliberately narrow generated task families. The generated setting improves experimental control but limits conclusions about natural images, while visible explanations should not be treated as hidden chain-of-thought.

\vfill
Future work should evaluate broader proprietary and open-weight model coverage, extend the generator to new spatial relations, and test \statesightsteps through prompting and training ablations. Candidate extensions include path structure, symmetry completion, layered alignment, rotations, clutter, and more complex occlusion.

\vfill
Broader external validation should connect the generated tasks to natural-image spatial benchmarks. SpatialSense and Visual Spatial Reasoning test object relations under natural variation, What's Up isolates changes in relative position, and SugarCrepe examines compositional benchmark shortcuts \cite{yang2019spatialsense,liu2022vsr,kamath2023whatsup,hsieh2023sugarcrepe}. These comparisons would clarify which \statesight failures persist beyond the controlled rendering domain.
\end{minipage}

\clearpage

\begin{center}

\noindent
\textbf{Table 2: Uncertainty summaries for the 300-item direct evaluations.}

\par\vspace{0.25em}

\begin{minipage}[t]{0.56\textwidth}
\centering
\small
\setlength{\tabcolsep}{5pt}
\renewcommand{\arraystretch}{1.08}

\begin{tabular}{@{}llcc@{}}
\toprule
Task & Model & Accuracy & Wilson 95\% CI \\
\midrule
Cube nets & GPT-5.5 & 59.3\% & [53.7, 64.7] \\
Cube nets & Claude Sonnet 5 & 53.3\% & [47.7, 58.9] \\
Cube towers & GPT-5.5 & 33.3\% & [28.2, 38.8] \\
Cube towers & Claude Sonnet 5 & 18.7\% & [14.7, 23.5] \\
Connected grids & GPT-5.5 & 28.3\% & [23.5, 33.7] \\
Connected grids & Claude Sonnet 5 & 7.3\% & [4.9, 10.9] \\
\bottomrule
\end{tabular}
\end{minipage}
\hfill
\begin{minipage}[t]{0.40\textwidth}
\centering
\small
\setlength{\tabcolsep}{4pt}
\renewcommand{\arraystretch}{1.08}

\begin{tabular}{@{}lcc@{}}
\toprule
Task & GPT--Claude & Paired 95\% CI \\
\midrule
Cube nets & +6.0 & [0.7, 11.3] \\
Cube towers & +14.7 & [10.3, 19.0] \\
Connected grids & +21.0 & [15.7, 26.3] \\
\bottomrule
\end{tabular}
\end{minipage}

\end{center}

\par\vspace{0.12em}

\noindent
Wilson intervals summarize uncertainty in each direct accuracy estimate.
The paired intervals resample shared item identifiers and exclude zero
on all three tasks, supporting the observed ordering of the two
evaluated models.

\vspace{0.55em}
\noindent
\begin{minipage}[t][0.68\textheight][s]{0.46\textwidth}
\vspace{0pt}
\centering
\includegraphics[width=0.96\linewidth,keepaspectratio]{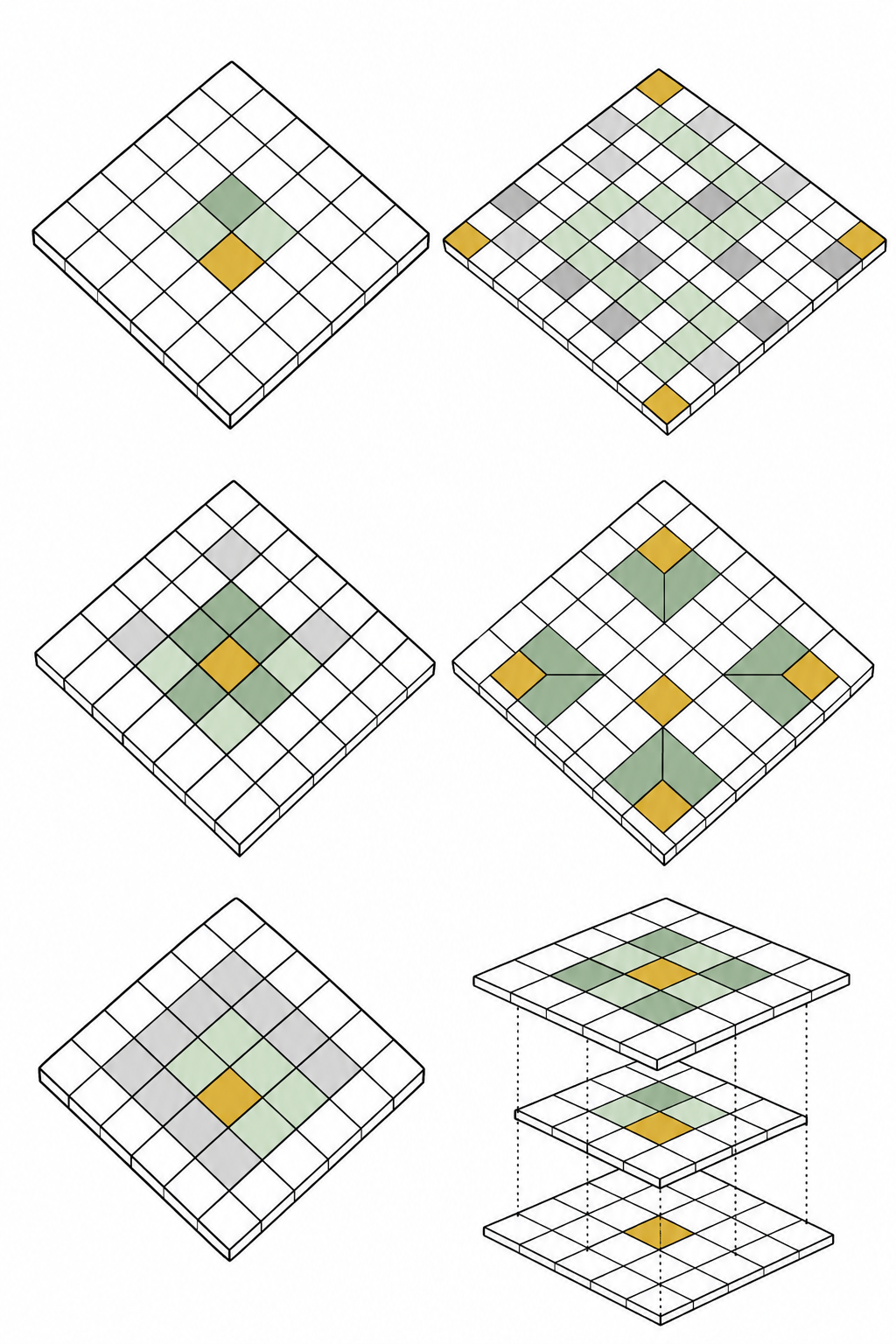}
\refstepcounter{figure}
\label{fig:future}
\par\vspace{0.12em}
\noindent\textbf{Figure \thefigure: Candidate future spatial-state task families.}
\par\vspace{0.06em}
\raggedright
The six panels show conceptual path-structure, symmetry-completion, and layered-alignment variants. None has yet been implemented or evaluated, but each could retain a symbolic state and deterministic oracle label.

\par\vspace{0.18em}
\noindent
A public release should include the generators, rendered data, prompts, truth labels, evaluation scripts, documentation, licensing, and machine-readable metadata. This would support comparison under one exact-match protocol and separate model improvements from changes in prompting or scoring.
\end{minipage}
\hfill
\begin{minipage}[t][0.68\textheight][s]{0.50\textwidth}

\section*{Acknowledgments}

The author thanks Dr.\ Mingrui Liu, Rui Yu, and the George Mason University Aspiring Scientists Summer Internship Program (ASSIP) for the opportunity to conduct this research and for their invaluable support throughout the project.

\vspace{0.35em}
\end{minipage}

\end{document}